\documentclass[runningheads]{llncs}
\usepackage{graphicx}
\usepackage{graphicx}
\usepackage{amsmath}
\usepackage{amssymb}
\usepackage{booktabs}
\usepackage{xcolor}
\usepackage{bm}
\usepackage{pifont}
\usepackage{enumitem}
\usepackage{soul}
\usepackage{dblfloatfix}
\usepackage{siunitx}
\usepackage{balance}
\usepackage[pagebackref,breaklinks,colorlinks]{hyperref}
\usepackage{url}
\usepackage{multirow}
\usepackage{xspace}
\usepackage{rotating}
\usepackage{framed}
\usepackage{bbm}
\usepackage{gensymb}
\usepackage{makecell}
\usepackage[caption=false]{subfig}
\usepackage[export]{adjustbox}
\usepackage{soul}

\def \ie {\emph{i.e.},}
\def \eg {\emph{e.g.},}

\newcommand{\tit}[1]{\smallbreak\noindent\textbf{#1.}}

\begin{document}
\title{How\hspace{0.117cm}to\hspace{0.117cm}Choose\hspace{0.117cm}Pretrained\hspace{0.117cm}Handwriting\hspace{0.117cm}Recognition\hspace{0.117cm}Models\hspace{0.117cm}for\hspace{0.117cm}Single\hspace{0.117cm}Writer\hspace{0.117cm}Fine-Tuning}

\titlerunning{Pre-trained HTR models for single writer fine-tuning}

\author{Vittorio Pippi\inst{1}\orcidID{0009-0001-7365-6348} \and Silvia Cascianelli\inst{1}\orcidID{0000-0001-7885-6050} \and Christopher Kermorvant\inst{2}\inst{3}\orcidID{0000-0002-7508-4080} \and \\ Rita Cucchiara\inst{1}\orcidID{0000-0002-2239-283X}}

\authorrunning{V. Pippi et al.}

\institute{University of Modena and Reggio Emilia, Modena, Italy \and TEKLIA, Paris, France \and LITIS, Université de Rouen - Normandie, France \email{\{vittorio.pippi,~silvia.cascianelli,~rita.cucchiara\}@unimore.it kermorvant@teklia.com}}

\maketitle              

\begin{abstract}
Recent advancements in Deep Learning-based Handwritten Text Recognition (HTR) have led to models with remarkable performance on both modern and historical manuscripts in large benchmark datasets. Nonetheless, those models struggle to obtain the same performance when applied to manuscripts with peculiar characteristics, such as language, paper support, ink, and author handwriting. This issue is very relevant for valuable but small collections of documents preserved in historical archives, for which obtaining sufficient annotated training data is costly or, in some cases, unfeasible. To overcome this challenge, a possible solution is to pretrain HTR models on large datasets and then fine-tune them on small single-author collections. In this paper, we take into account large, real benchmark datasets and synthetic ones obtained with a styled Handwritten Text Generation model. Through extensive experimental analysis, also considering the amount of fine-tuning lines, we give a quantitative indication of the most relevant characteristics of such data for obtaining an HTR model able to effectively transcribe manuscripts in small collections with as little as five real fine-tuning lines.

\end{abstract}\vspace{-2em}

\section{Introduction}
Digitization is becoming a crucial step for the efficient management, preservation, and valorization of documents, both in the cultural and industrial domains. For this reason, Document Analysis (DA) techniques, especially those intended to tackle challenging scenarios of handwritten text, are receiving significant interest from the research community. 
State-of-the-art Handwritten Text Recognition (HTR) models, trained on large publicly available datasets, can achieve impressive results when applied to documents with characteristics similar to those used during training. However, their performance is unsatisfactory when the data of the domain of interest are too different from the training ones.
In this respect, the small but valuable collections of historical manuscripts preserved in many archives pose a challenge for modern HTR models. In fact, such archives often contain few sample pages written by a specific but relevant author, with peculiar characteristics, both visual and linguistic. Thus, a strategy to obtain high HTR performance also for those documents is key to enabling the efficient digitization of such documents. 
A popular approach to deal with this scenario consists in pretraining the HTR model on large datasets, either real or synthetic, and then fine-tuning on a limited number of real data from the target domain. This strategy has also the potential to enable high-quality on-demand transcriptions of entire single-author collections, which might be a service of interest to the users of digital libraries and archives. In particular, the libraries can store pretrained HTR models, and users can request the transcription of the collection they are interested in by simply providing the annotation for a few lines (\eg~$5-15$). At that point, the most suitable pretrained model can be chosen based on the collection characteristics (\eg~language, period, authorship, style) possibly available in the form of metadata and fine-tuned on the user-provided annotations in a limited amount of time. Afterward, the fine-tuned model can transcribe the entire collection with low error. This kind of interaction with the collection can also benefit the overall experience of digital archives users. Note that, so far, interactive transcription enhancement has been explored in terms of language model refinement~\cite{martin2012multimodal}. With this work, we aim to explore a more holistic approach taking into account both language and appearance.

Note that, in literature, attempts have been made toward the use of synthetic data for pretraining HTR models~\cite{granet2018transfer,aradillas2020boosting,kang2022pay,wick2021rescoring,li2021trocr}. These strategies are as effective as more similar the synthetic data are to the real ones~\cite{cascianelli2021learning}. In this line, Handwritten Text Generation (HTG) techniques are emerging~\cite{bhunia2021handwriting,bhunia2021metahtr,fogel2020scrabblegan,kang2020ganwriting}, especially styled HTG ones, which might allow generating training data with the characteristics needed for HTR on specific domains. In fact, models for styled HTG can produce images with arbitrary text in the desired handwriting starting from a few style example images. These models often comprise an encoder to obtain writer-specific style features and a generator, which is fed with the style features and content tokens representing the characters to produce text images conditioned on the desired style and content. 
In light of this, in this paper, we consider pretraining plus fine-tuning on an automatically generated author-specific synthetic dataset, which is obtained by exploiting a State-of-the-Art styled HTG network. Moreover, we evaluate pretraining on existing benchmark datasets of various languages, with a varied number of authors and of various periods. 
This way, we investigate whether it is feasible to obtain an effective pipeline for interactive, on-demand HTR of single-author collections. In particular, we provide a set of quantitative guidelines taking into account both visual and linguistic aspects, for designing the most effective pipeline for an HTR model able to transcribe specific manuscript collections with low error after fine-tuning on as little as $5$ lines from the target manuscript. 
Potentially, the defined guidelines for choosing the most suitable pretrained model can either be exploited by the archive management or presented to the user who would be more involved in the transcription process.

\section{Related Work} \label{sec:related}
\tit{Strategies for HTR}
Due to its practical interest in both industrial and cultural domain applications, HTR is a widely-investigated research topic. Despite that, it remains a challenging task. HTR can be performed on single characters, which is a popular choice in the case of idiomatic languages~\cite{cilia2019ranking}, single words~\cite{such2018fully,bhunia2019handwriting}, or entire lines~\cite{shi2016end,puigcerver2017multidimensional}, paragraphs, and pages~\cite{moysset2017full,bluche2017scan,bluche2016joint,wigington2018start,clanuwat2019kuronet}. The line-level variant is one of the most popular for non-idiomatic language, both standalone and as part of a page-level system~\cite{yousef2020origaminet,moysset2017full,bluche2017scan}
The most used learning-based solutions for HTR rely on Multi-Dimensional Long Short-Term Memory networks (MD-LSTMs) ~\cite{graves2009offline} or on the combination of convolutional and one-dimensional LSTMs~\cite{shi2016end,puigcerver2017multidimensional,pham2014dropout,voigtlaender2016handwriting,bluche2017gated} to represent the text image and on the Connectionist Temporal Classifier (CTC) decoding strategy to output the transcription~\cite{graves2009offline,bluche2016joint}. Alternatively to approaches exploiting recurrent models, fully-convolutional networks have been proposed for HTR~\cite{yousef2020origaminet,coquenet2020recurrence}, as well as solutions~\cite{kang2022pay,li2021trocr,wick2021transformer} based on Transformer encoder-decoder architectures~\cite{vaswani2017attention}.
Finally, it is worth noting that explicit language models or lexicons can be exploited to refine the transcription. However, this strategy is all the more effective the more the language of the transcribed images is regular (\ie~it contains no errors, uncommon words, and proper nouns) and well-represented. For this reason, employing language models is not always feasible, especially when dealing with historical manuscripts.

\tit{Strategies for HTG}
HTG is an increasingly popular research area aimed at producing realistic images of handwritten text. In the styled variant of the task, which we consider in this work, the goal is to generate writer-specific handwritten text images from just a few example images of the writer's style to mimic~\cite{bhunia2021handwriting,fogel2020scrabblegan,kang2020ganwriting}. 
The early approaches to HTG, either styled or not, were able to obtain impressive results, but at the cost of heavy human intervention and feature handcrafting~\cite{wang2005combining,haines2016my}. Recently-proposed learning-based solutions, instead, are fully automatic. Usually, these strategies entail using
generative adversarial networks (GANs)~\cite{goodfellow2014generative}. In the case of non-styled HTG, these can be unconditioned~\cite{alonso2019adversarial,fogel2020scrabblegan}. For styled HTG, instead, the employed GANs are conditioned on style features extracted by an encoder from the handwriting style sample images. Note that the style examples can be line images~\cite{davis2020text}, a few images of words~\cite{kang2020ganwriting,bhunia2021handwriting}, or a single image~\cite{mattick2021smartpatch}. 
It is also worth mentioning a more recently-proposed approach based on an encoder-decoder generative Transformer~\cite{bhunia2021handwriting}.

\tit{Synthetic data for HTR}
Lack of training data is a major challenge in HTR, especially in the case of single-author documents or ancient manuscripts that exhibit peculiar characteristics. A possible strategy to tackle this issue is to perform data augmentation either in terms of generic color modifications and geometric distortions~\cite{voigtlaender2016handwriting,puigcerver2017multidimensional,wigington2017data} or image modifications carefully designed to match the characteristics of the target data~\cite{chammas2018handwriting}.
Another popular strategy entails pretraining the HTR model on large datasets and then fine-tuning it on the target data~\cite{granet2018transfer,jaramillo2018boosting,soullard2019improving}, which has been proven to be more beneficial than data augmentation for historical manuscripts~\cite{aradillas2020boosting}. The pretraining dataset can be real (\eg~a publicly available benchmark dataset) or synthetic, generally obtained by altering images of text rendered in calligraphic fonts~\cite{shen2016method,kang2022pay}.
In an attempt to generate more realistic-looking text images, some recent works exploited HTG models, either styled or not, to generate synthetic data for training HTR models and boost their performance on real data. For example, in~\cite{souibgui2022one}, the authors exploited a compositional approach based on Bayesian Program Learning to generate the symbols in a ciphered corpus and then combined them into realistic-looking text lines for training an HTR model to transcribe historical ciphered manuscripts. The benefits of training HTR models on generated text lines in various styles have been investigated also in~\cite{kang2021content}, where the authors applied a styled HTG model to obtain a pretraining dataset. Finally, as for the single-author scenario, in~\cite{cascianelli2021learning}, the authors showed the benefits of pretraining the HTR model on synthetic data that faithfully resemble the real ones over pretraining on generic various-styles images. Their approach, however, heavily relies on human effort to obtain such high-quality synthetic data. In sight of this, in this work, we focus on synthetic data obtained from a fully-automatic HTG model.

\section{Proposed Approach} \label{sec:method}
\begin{figure*}[t]
    \centering
    \includegraphics[width=0.9\textwidth]{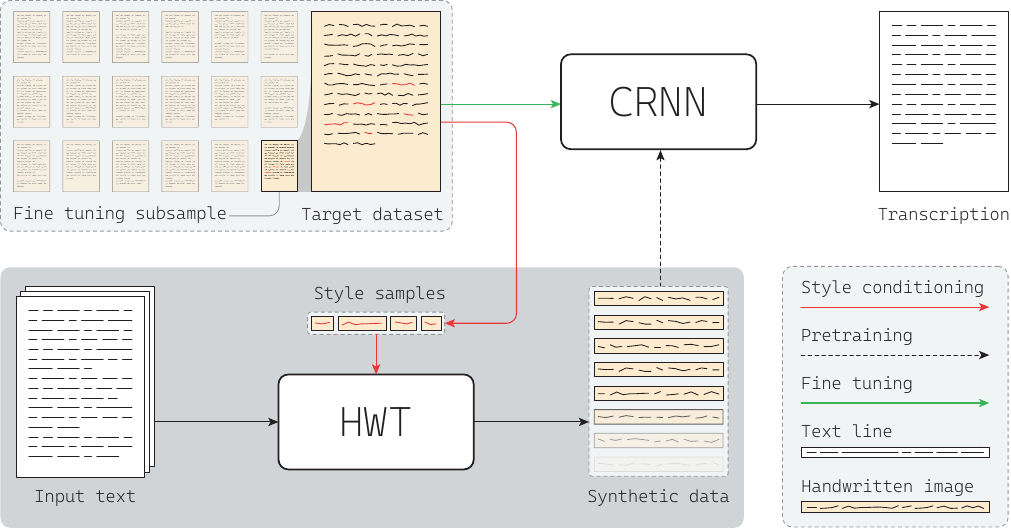}
    \caption{Overview of our pipeline for synthetic data generation from collection-specific handwritten lines. The generation process renders handwritten line images from a given text conditioned by a few style samples from the target dataset. Then, the synthetic dataset is used to pretrain the CRNN.}\vspace{-1em}
    \label{fig:overview}
\end{figure*}

In this work, we explore a pipeline for obtaining good-quality line-level transcriptions of manuscript collections with peculiar characteristics (in terms of handwriting, language, and paper support) by exploiting pretraining on large datasets and fine-tuning on a small amount of samples from the target collection. In particular, we consider pretraining on real datasets and on synthetic ones obtained via a few-shot styled HTG model to better reflect the characteristics of the target data.
To build the synthetic datasets, we need a few images of words (15, in this work, as in~\cite{kang2020ganwriting,bhunia2021handwriting}) that can be easily obtained from digitized manuscripts in the small collection of interest. Moreover, we need to specify the text to be rendered in the desired style. In this work, we consider two typical scenarios in digital libraries and archives: one in which only the language of the target manuscript is known and one in which also the information about the author is available. 
If the author of the collection of interest is known and there exist some other transcribed texts by the same author, we propose to make the HTG model generate these texts. In case only the language is known (or if no other texts by the same author are available), we make the HTG model generate texts in the same language as the target collection. Note that, in both cases, the HTG model outputs images of handwritten words, which we then combine into lines of varying length. 
In the following, we describe the HTR network used for the transcription and the HTG network used to generate the synthetic pretraining data. An overview of our complete pipeline is depicted in Figure~\ref{fig:overview}.

\subsection{HTR Model}
Combining convolutional neural networks and recurrent neural networks for HTR has been the standard choice for years, and many currently available transcription services feature this kind of models for their efficiency. In this work, we consider a model featuring one-dimensional LSTMs, since these have been proven to be comparable or superior to MD-LSTMs~\cite{graves2009offline} and are faster to train~\cite{puigcerver2017multidimensional}.

In particular, in this work, we consider a variant of the approach proposed in~\cite{shi2016end} (referred to as \textbf{CRNN} in the following). The convolutional part of the architecture features seven convolutional blocks. For the first six blocks, we adopt the same architecture as in the VGG-11 network, with the difference of applying rectangular pooling in the last two max-pooling layers to better reflect the aspect ratio of text lines images. The seventh convolutional block has a $2\times2$ kernel. In the adopted variant, the convolutional component features Deformable Convolutions~\cite{dai2017deformable}, as proposed in~\cite{cojocaru2020watch,cascianelli2021learning,cascianelli2022boosting}, which enhances the performance. 
The feature map of the last convolutional layer is a  $2\times W \times 512$ tensor, where $W$ depends on the width of the input text line image. This tensor is collapsed along the channel dimension to obtain a sequence of $W$ feature vectors of $1024$ elements, which is fed to the recurrent part of the architecture. 
This consists of two  Bidirectional LSTM layers with $512$ hidden units each, separated by a dropout layer with probability $0.5$. The recurrent part outputs the probability of each feature vector in the sequence to contain each of the characters in a charset. 

As customary in HTR, the model is trained to optimize the CTC loss, and thus, a special \emph{blank} character is included in the model charset.
Note that we do not use any language model in combination with the HTR network to achieve cross-language adaptability.

\subsection{HTG Approach}
Styled HTG models allow to efficiently obtain a large number of synthetic text images in the handwriting of the desired author, which can then be used to train an HTR model tailored for the author of interest. In this work, we build upon the transformer-based few-shot styled HTG model recently proposed in~\cite{bhunia2021handwriting}, namely Handwriting Transformer (\textbf{HWT}). 

The HTG approach applied is a Convolutional-Transformer encoder-decoder architecture. The handwriting examples are first fed to a convolutional feature extractor in the encoder (namely, a ResNet18), whose outputs are passed through a multi-layer, multi-headed Transformer encoder (with $3$ layers and $8$ attention heads) and thus enriched with long-range dependencies thanks to the self-attention mechanism. 
The resulting style vectors are used as keys and values in a multi-layer, multi-headed Transformer decoder (with $3$ layers and $8$ attention heads as the encoder) that performs cross-attention with vectors representing the characters in the words to be rendered. Normal gaussian noise is added to the resulting vectors to obtain some variability, and those are then fed into a convolutional decoder made of four residual blocks and a $\mathrm{tanh}$ activation, which outputs the final styled word images.

The HWT model is trained alongside additional blocks by optimizing a multiple-term loss function. In particular, we follow the adversarial paradigm with the hinge adversarial loss~\cite{lim2017geometric} and train HWT together with a convolutional discriminator. Moreover, to enforce the generation of readable word images, we include a CTC loss term obtained by making an HTR model predict the textual content in the generated images. Also this HTR model is inspired by the architecture proposed in~\cite{shi2016end}. Finally, to 
force HWT to faithfully render the desired style, we use two additional loss terms. One is the cross-entropy loss of a convolutional classifier aimed at classifying the generated images based on the writers in the HWT training set. The other is a cycle consistency loss term given by the $l_1$-norm of the difference between the encodings of the real and the generated images obtained by the encoder part of HWT.

\section{Experiments}\label{sec:experiments}

In this section, we describe our experimental analysis. First, we give further implementation details on the adopted HTR and HTG models. Then, we describe the considered small, single-author target datasets, the real benchmark datasets (whose details are reported in Table~\ref{tab:datasets}, and some samples in Figure~\ref{fig:datasets_samples}), and the details on the procedure to build the synthetic pretraining datasets. Finally, we describe the evaluation protocol applied and discuss the results obtained.

\subsection{Implementation Details}
\label{sec:implementation_details}

The experiments for this work, both the HTG and HTR part, have been performed on a single 
NVIDIA RTX 2080~Ti GPU.

\tit{CRNN}
To train the CRNN model, we rescale all images to a height of 60 pixels, maintaining the original aspect ratio, and then normalize them between $-1$ and $1$.
Additionally, when pretraining, we apply the following augmentations. We modify the brightness of the image with a factor randomly chosen between $0.5$ and $5$, the contrast with a factor randomly selected from between $0.1$ and $10$, the saturation with a factor randomly chosen between $0$ and $5$, and the hue with a factor randomly selected from between $-0.1$ and $0.1$. Moreover, we apply Gaussian blur whose kernel size is set to $5$, and the standard deviation is randomly chosen between $0.1$ and $2$. Finally, we apply a geometric distortion chosen among the following: random rotation (between $-1\degree$ and $1\degree$), affine transformation (with random rotation between $-1\degree$ and $1\degree$ and random shear between $-50\degree$ and $30\degree$), and random homography. 
We use a batch size of $16$ when pretraining and a batch size of $8$ when fine-tuning and training from scratch. All experiments use a learning rate of $10^{-4}$.
We train the proposed model with Adam as optimizer, with $\beta_1 = 0.9$ and $\beta_2 = 0.999$, and a scheduler to reduce the learning rate by $10\%$ if the model reaches a plateau for the CER on the validation set.
We train the models with a patience of $20$ epochs for the CER on the validation set. Note that when fine-tuning, we usually obtain the best CER within the second epoch, which takes roughly less than an hour.

\tit{HWT}
We train the HWT model on the IAM dataset in the word-level setting (See~\ref{ssec:real_datasets}). All image samples are grayscale images resized to a height equal to $32$ pixels, maintaining the same aspect ratio and normalized between $-1$ and $1$. We train the model with the same settings as those used in the original paper~\cite{bhunia2021handwriting} for $7000$ epochs. The HWT model generates single-word images whose characters occupy, on average, $16$ pixels each. For this reason, we concatenate different images to make a line with a spacing of $16$ pixels between the words.

\begin{table}[t]
\centering
\small
\setlength{\tabcolsep}{.3em}
\caption{Characteristics of the considered line-level datasets.\vspace{-3em}}
\label{tab:datasets}
\resizebox{\columnwidth}{!}{%
\begin{tabular}{lc ccccc}
\toprule 
 & & \textbf{Training Lines} & \textbf{Charset} & \textbf{Period} & \textbf{Language} & \textbf{Authors}\\
\midrule
Washington~\cite{fischer2012lexicon}        & & 526 & 68 & 1755 & English & One\footnotemark[1]\\
Saint Gall~\cite{fischer2011transcription}  & & 468 & 49 & ca 890-900 & Latin & One\\
Leopardi~\cite{cascianelli2021learning}     & & 1303 & 76 & 1818-1832 & Italian & One\\
\midrule
IAM~\cite{marti2002iam}                     & & 6482 & 79 & Modern & English & Many\\
ICFHR16~\cite{sanchez2016icfhr2016}         & & 8367 & 88 & 1470-1805 & German & Many\\
Rodrigo~\cite{serrano2010rodrigo}           & & 9000 & 105 & 1545 & Spanish & One\\
ICFHR14~\cite{sanchez2014icfhr2014}         & & 9198 & 93 & ca 1760-1832 & English & One\\
RIMES~\cite{augustin2006rimes}              & & 10188 & 95 & Modern & French & Many\\
NorHand~\cite{maarand2022comprehensive} & & 19653 & 111 & 1820-1940 & Norwegian & Many\\
LAM~\cite{cascianelli2022lam}               & & 19830 & 89 & 1691-1750 & Italian & One\\
\midrule
Synthetic for Washington                    & & 23121 & 78 & - & English & One\\
Synthetic for Saint Gall                    & & 70494 & 66 & - & Latin & One\\
Synthetic for Leopardi                      & & 89068 & 113 & - & Italian & One\\
\bottomrule
\addlinespace[0.1cm]
\multicolumn{7}{l}{\footnotesize 1 -  A small number of lines are by another writer.}
\end{tabular}
}\\
\vspace{-.3cm}
\end{table}

\subsection{Target Datasets}
\tit{Leopardi~\cite{cascianelli2021learning}} The Leopardi dataset consists of a small collection of early 19\textsuperscript{th} Century letters written in Italian by the Romanticism philologist, writer, and poet Giacomo Leopardi. It contains 1303 training lines, 596 validation lines, and 587 test lines. All samples are RGB scans of documents written with ink on ancient paper.

\tit{Washington~\cite{fischer2012lexicon}} The George Washington dataset contains 20 English letters written by American President George Washington and one of his collaborators in 1755. The dataset is divided into 526 training lines, 65 validation lines, and 65 test lines. All samples are binary images.

\tit{Saint Gall~\cite{fischer2011transcription}} The Saint Gall dataset contains 60 pages of a handwritten historical manuscript written in Latin by a single author at the end of the 9\textsuperscript{th} Century. The dataset is divided into 468 training lines, 235 validation lines, and 707 test lines. All samples are binary images. 

\subsection{Pretraining Real Datasets}\label{ssec:real_datasets}
\begin{figure*}[h!]
    \centering
    \setlength{\tabcolsep}{.2em}
    \begin{tabular}{c}
    \midrule
    \multicolumn{1}{l}{Target Datasets} \\ \\
    \begin{tabular}{c c c}
        \includegraphics[height=0.55cm,width=4cm]{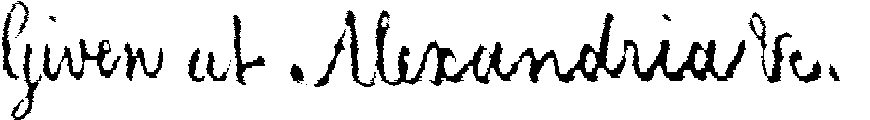} &
        \includegraphics[height=0.55cm,width=4cm]{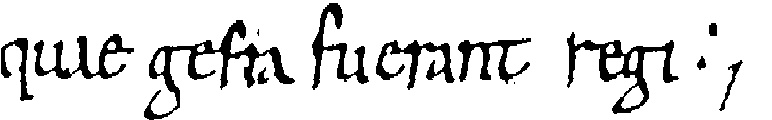} &  
        \includegraphics[height=0.55cm,width=4cm]{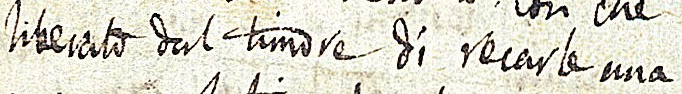} 
        \\
        \includegraphics[height=0.55cm,width=4cm]{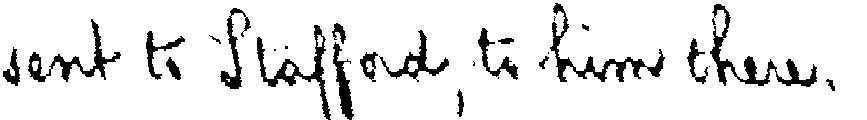} &
        \includegraphics[height=0.55cm,width=4cm]{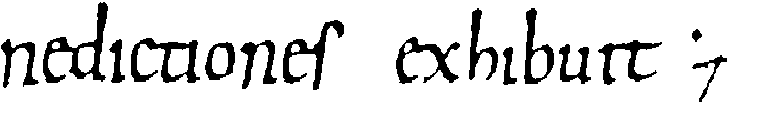} &  
        \includegraphics[height=0.55cm,width=4cm]{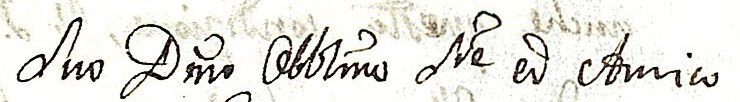} 
        \\
        \includegraphics[height=0.55cm,width=4cm]{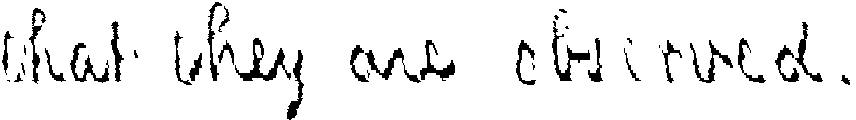} &
        \includegraphics[height=0.55cm,width=4cm]{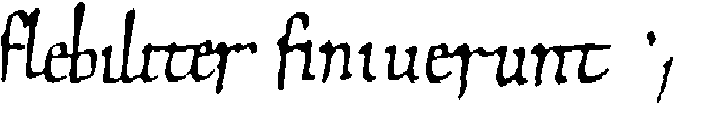} &  
        \includegraphics[height=0.55cm,width=4cm]{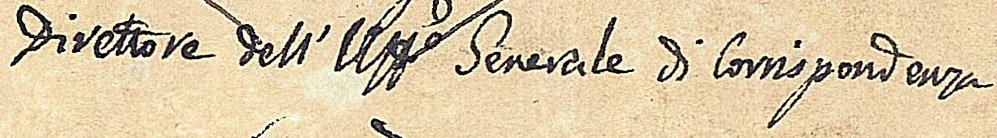} 
        \\
        \textbf{Washington} & \textbf{Saint Gall} & \textbf{Leopardi}
    \end{tabular}
    \\ \\
    \midrule
    \multicolumn{1}{l}{Pretraining Datasets} \\ \\
    \begin{tabular}{c c}
        \includegraphics[height=0.55cm,width=4cm]{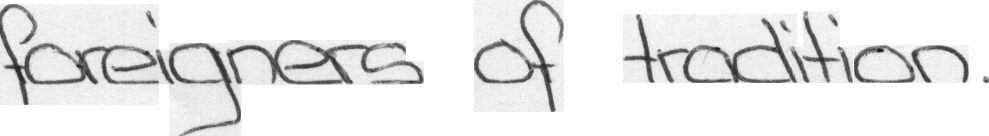} &  
        \includegraphics[height=0.55cm,width=4cm]{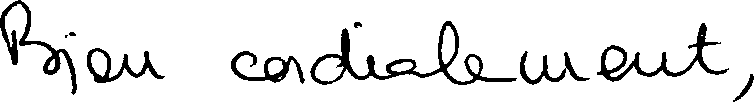} 
        \\
        \includegraphics[height=0.55cm,width=4cm]{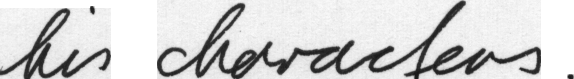} &  
        \includegraphics[height=0.55cm,width=4cm]{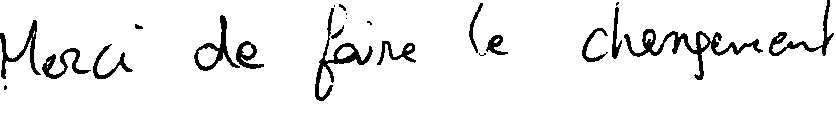} 
        \\
        \includegraphics[height=0.55cm,width=4cm]{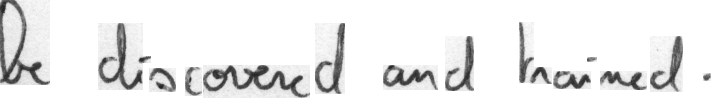} & 
        \includegraphics[height=0.55cm,width=4cm]{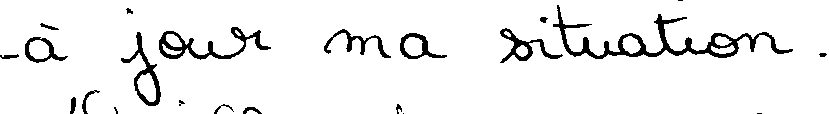} 
        \\
        \textbf{IAM} & \textbf{RIMES} 
    \end{tabular}
    \\ \\
    \begin{tabular}{c c}
        \includegraphics[height=0.55cm,width=4cm]{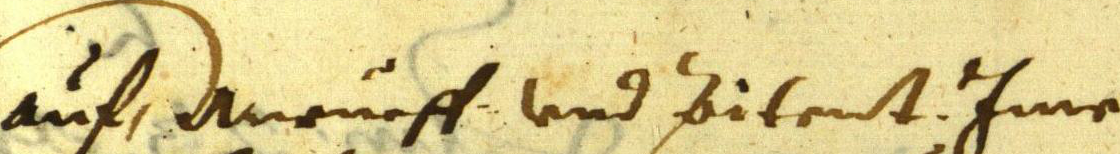} & 
        \includegraphics[height=0.55cm,width=4cm]{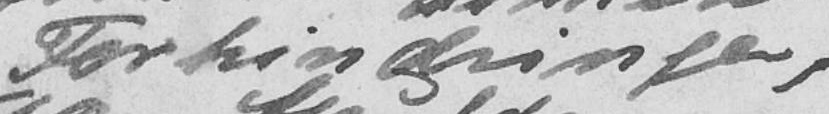} 
        \\
        \includegraphics[height=0.55cm,width=4cm]{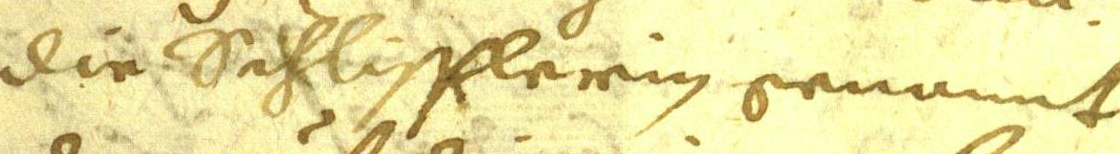} & 
        \includegraphics[height=0.55cm,width=4cm]{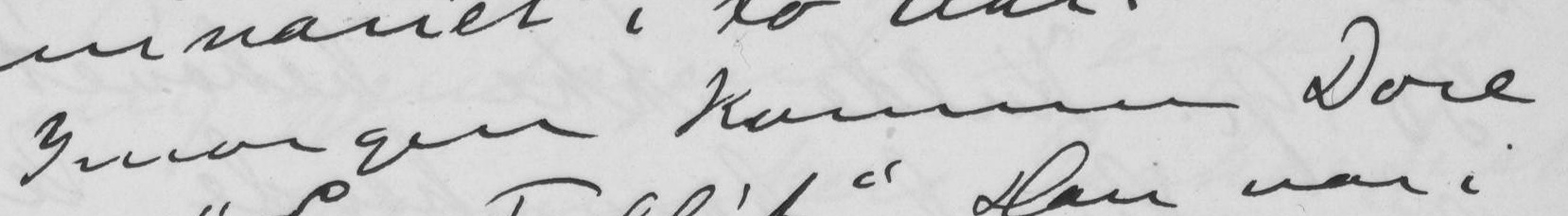} 
        \\
        \includegraphics[height=0.55cm,width=4cm]{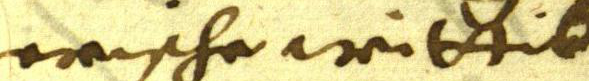} & 
        \includegraphics[height=0.55cm,width=4cm]{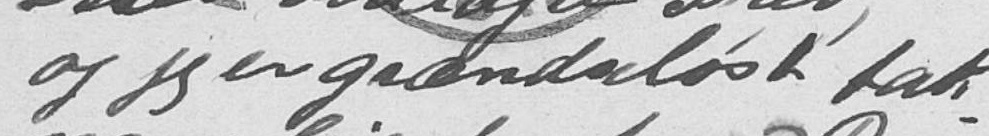} 
        \\
        \textbf{ICFHR16} & \textbf{NorHand}
    \end{tabular}
    \\ \\
    \begin{tabular}{c c c}
        \includegraphics[height=0.55cm,width=4cm]{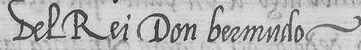} &
        \includegraphics[height=0.55cm,width=4cm]{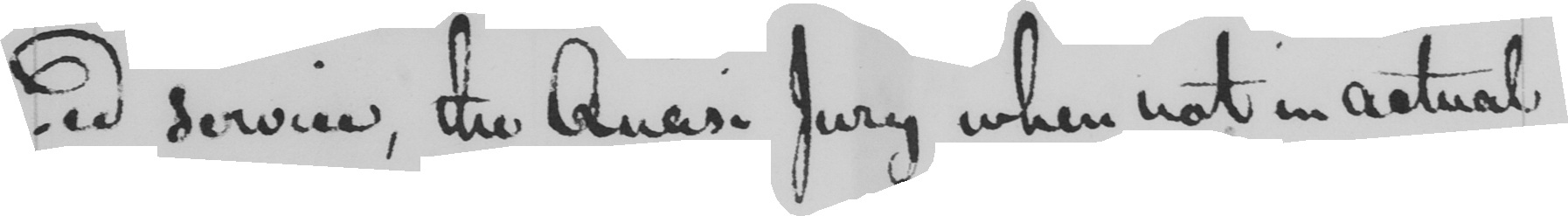} &  
        \includegraphics[height=0.55cm,width=4cm]{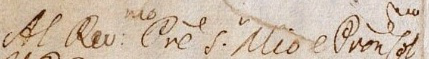} 
        \\
        \includegraphics[height=0.55cm,width=4cm]{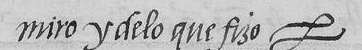} &
        \includegraphics[height=0.55cm,width=4cm]{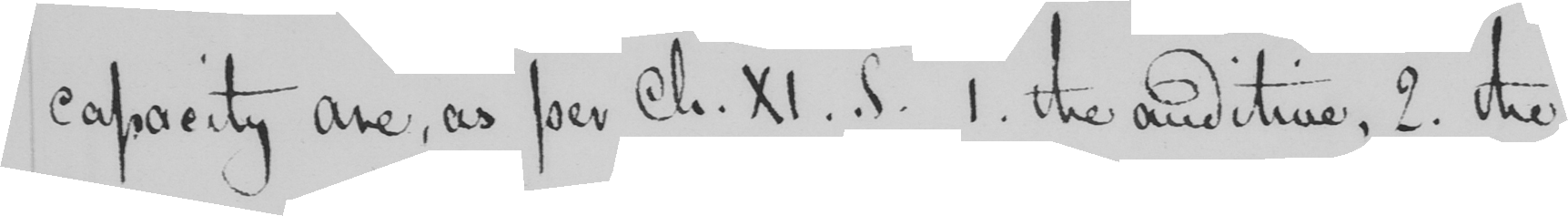} &  
        \includegraphics[height=0.55cm,width=4cm]{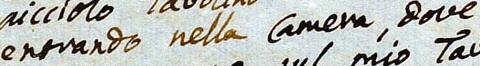} 
        \\
        \includegraphics[height=0.55cm,width=4cm]{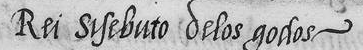} &
        \includegraphics[height=0.55cm,width=4cm]{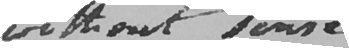} &  
        \includegraphics[height=0.55cm,width=4cm]{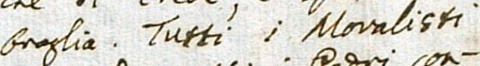} 
        \\
        \textbf{Rodrigo} & \textbf{ICFHR14} & \textbf{LAM}
    \end{tabular}
    \end{tabular}
        \caption{Exemplar line images from the considered real datasets. These datasets are in different languages, from different periods, and with varied number of writers.}
    \label{fig:datasets_samples}
    \vspace{-0.3cm}
\end{figure*}
\tit{Rodrigo~\cite{serrano2010rodrigo}} The Rodrigo dataset contains 853 pages written in Spanish by a single author. The pages come from a manuscript entitled “Historia de Espa\~{n}a del ar\c{c}obispo Don Rodrigo,” written in 1545. All samples are grayscale scans of documents written with ink on ancient paper.

\tit{NorHand~\cite{maarand2022comprehensive}} The NorHand dataset consists of 4144 pages, mainly of diaries and letters written by 15 Norwegian authors from approximately 1820 to 1940. The training set includes 19653 lines. All samples are grayscale scans of documents written with ink on yellowed paper. 

\tit{LAM~\cite{cascianelli2022lam}} The LAM dataset contains 1171 letters by the historian Ludovico Antonio Muratori in Italian from 1691 to 1750, and thus, exhibits a certain degree of variability due to this wide time-span. The training set includes 19830 lines. All samples are RGB scans of documents written with ink on ancient paper.

\tit{ICFHR14~\cite{sanchez2014icfhr2014}} The ICFHR14 dataset contains a collection of 433 pages on law and moral philosophy written by the English philosopher Jeremy Bentham from 1760 to 1832. The training set includes 9198 lines. All samples are grayscale scans of documents written with ink on yellowed paper.

\tit{ICFHR16~\cite{sanchez2016icfhr2016}} The ICFHR16 dataset consists of a subset of 400 pages from the Ratsprotokolle collection written from 1470 to 1805 in Early Modern German. The number of writers is unknown. The training set includes 8367 lines. All samples are grayscale scans of documents written with ink on yellowed paper.

\tit{RIMES~\cite{augustin2006rimes}} The RIMES database (Reconnaissance et Indexation de données Manuscrites et de fac similÉS / Recognition and Indexing of handwritten documents and faxes) consists of 12723 pages of scanned letters written in French by 1300 different authors. The training set includes 10188 lines. All samples are binary images. 

\tit{IAM~\cite{marti2002iam}} The IAM Handwriting Database 3.0 is a modern collection of 1539 scanned pages written in English by 657 different authors. The IAM dataset comes in different settings: we use the word-level setting to train the HTG network and the line-level setting for pretraining the HTR network. In particular, we train the HTG network with all the available words from 339 different authors in the word-level training set. For the line-level setting, there are 6482 lines for training, 976 for validation, and 2915 for test. All samples are grayscale scans of documents written with ink on white paper and cleaned digitally.

\subsection{Synthetic Data}
We use the HWT model, trained on the word-level IAM dataset, to generate synthetic data specific to each of the target datasets. In particular, for each dataset, we isolate a number of word images and repeatedly choose randomly $15$ of them to serve as style examples. As for the textual content, we use some of Giacomo Leopardi's proses for the Leopardi dataset, some of George Washington's diaries for the Washington dataset, and a Bible in medieval Latin for the Saint Gall dataset. In this way, the language of the synthetic datasets more closely resembles that of the target datasets. In the following, we refer to these synthetic datasets as \textbf{HWT-Generated}. Note that HTW outputs words with characters whose average width is $16$ pixels. In our experiments, we also consider a variant of the synthetic datasets in which the generated images are resized in width to match the average character width of the target dataset. The resulting datasets are referred to as \textbf{HWT-Generated+WA} (Width Adjustment - WA), in the following. As a final variant, we generate synthetic datasets with the same textual content as the HWT-Generated versions but with style images from the IAM dataset. This variant is referred to as \textbf{HWT-Generated+WA+VS} (Varied Styles - VS), in the following.
Additionally, for the Leopardi dataset, we exploit the synthetic text lines released alongside the dataset. These have been obtained by rendering the same text that we use for the HWT-Generated Leopardi with a manually built randomized font mimicking the author's handwriting. In the following, we refer to this synthetic dataset as \textbf{Human-Synthesized}.

\subsection{Evaluation Protocol}
To evaluate the effect of the pretraining and fine-tuning strategy in the scenario in which only a few lines in the target dataset are annotated, we perform fine-tuning on a progressively smaller number of training lines, accounting for the $100\%$ (taken as a reference), $50\%$, $5\%$, $2.5\%$, and $1.25\%$, respectively. As an additional comparison, we train from scratch on the same amount of training lines. Moreover, we consider direct transfer of the pretrained models on the target datasets. The transcription performance is reported in terms of the commonly-used Character Error Rate (CER) and Word Error Rate (WER) scores.

To give further insights into the characteristics of the training data from a linguistic point of view, we calculate the Kullback Leiber Divergence between the distributions of character unigrams, bigrams, and trigrams in each target and pretraining dataset. Additionally, we report the Lexical Similarity~\cite{bella2021database} between the languages of the considered datasets. This quantifies the lexical similarity between pairs of languages based on words in both languages having a common origin and similar pronunciation and meaning.

\subsection{Results}
\begin{table}[t]
\footnotesize
\centering
\caption{Performance of the considered model when pretrained on real datasets or on differently-obtained synthetic datasets (WA stands for `width adjustment' and VS for `varied styles') and directly applied to the considered target datasets test set.}
\label{tab:direct_transfer}
\setlength{\tabcolsep}{.4em}
\resizebox{0.8\linewidth}{!}{
\begin{tabular}{lc cc c cc c cc}
\toprule
& & \multicolumn{2}{c}{\textbf{Leopardi}} & & \multicolumn{2}{c}{\textbf{Washington}} & & \multicolumn{2}{c}{\textbf{Saint Gall}} \\
\cmidrule{3-4} \cmidrule{6-7} \cmidrule{9-10} 
& & CER & WER & & CER & WER & & CER & WER\\
\midrule
IAM          & & 57.0 & 95.6  & & 50.5 & \textbf{88.7}  & & \textbf{49.9} & \textbf{97.7} \\
RIMES        & & 68.0 & 97.2  & & \textbf{48.2} & 96.2  & & 51.2 & 98.5 \\
Rodrigo      & & 88.6 & 99.5  & & 84.3 & 104.2 & & 55.2 & 111.4 \\
ICFHR14      & & 75.0 & 105.6 & & 83.6 & 128.8 & & 83.7 & 106.9 \\
ICFHR16      & & 80.5 & 104.5 & & 87.8 & 127.4 & & 80.1 & 106.3 \\
NorHand  & & 49.1 & 95.4  & & 64.9 & 101.4 & & 76.0 & 114.4 \\
LAM          & & \textbf{23.4} & \textbf{57.3}  & & 78.8 & 103.8 & & 78.0 & 99.5 \\
\midrule
Human-Synthesized    & & 56.9 & 95.3 & & -    & -     & & -    & -     \\ 
HWT-Generated        & & 93.2 & 99.4 & & 90.9 & 100.0 & & 83.7 & 106.9 \\ 
HWT-Generated+WA     & & 87.5 & 99.2 & & 87.5 & 100.0 & & 76.7 & 100.2 \\ 
HWT-Generated+WA+VS  & & 87.9 & 99.3 & & 83.0 & 98.6  & & 75.3 & 99.9  \\ 
\bottomrule
\end{tabular}
}
\vspace{-0.3cm}
\end{table}

As a baseline experiment, we perform direct transfer of the CRNN model pretrained on the real and synthetic data. The results are reported in Table~\ref{tab:direct_transfer}. It can be noticed that, on average, the models pretrained on the synthetic data perform worse than those trained on the real ones, except in the case of the Human-Synthesized dataset for Leopardi. In Tables~\ref{tab:leo_fine_tuning}-\ref{tab:saint_fine_tuning}, we report the results of fine-tuning on a few lines from the real datasets. These experiments aim to reflect the on-demand transcription application. As a reference, we also report the results of CRNN models trained from scratch on the same amount of lines. Note that when less than $230$ lines are used for training, the model did not converge, thus enforcing the need for pertaining. 

From the results, especially when fine-tuning on $1.25\%$ and $2.5\%$ of the training lines, where the most noticeable differences in performance appear, it emerges that for each target dataset, the best pretraining dataset can be identified, suggesting that the accurate selection of the pretraining data is key in boosting the recognition performance. In particular, the LAM dataset is overall the most suitable when working on the Leopardi dataset (see Table~\ref{tab:leo_fine_tuning}), the ICFHR14 is the most helpful when working on the Washington dataset (see Table~\ref{tab:wash_fine_tuning}), and Rodrigo when working on Saint Gall (see Table~\ref{tab:saint_fine_tuning}). 

Note that pretraining a network on a dataset different from the target one induces a bias that depends on some characteristics of the dataset used. In particular, the more the pertaining and target datasets are similar from the linguistic and visual point of view, the more useful the bias will be in terms of the resulting performance. 
In the following, we explore the main causes of this performance with the aim of tracing some guidelines for the selection of the most suitable pretraining dataset.

\tit{Appearance}
When choosing the dataset to be utilized during network pretraining, one aspect to consider is the overall visual appearance (\eg~paper support, ink color and thickness, and character width). To analyze the similarities, we refer to Figure~\ref{fig:datasets_samples}, showing some examples of the various datasets. It can be noticed that the samples in the LAM and the Leopardi datasets look similar, and this is reflected in the performance that CRNN reaches when pretrained on LAM and then fine-tuned on Leopardi. Conversely, there is a considerable difference between the samples in Leopardi and Rodrigo. As a result, pretraining on this latter dataset for HTR on Leopardi leads to poor performance.
A similar case can be made for the other two datasets, Saint Gall and Washington. For example, the images in Saint Gall have regular handwriting, similar to those in Rodrigo, which is one of the datasets that leads to the best performance. Moreover, the images in the Washington dataset are visually similar to those in RIMES or IAM. These two datasets are, in fact, those on which performing pretraining leads to the best performance on Washington. From these observations, we can conclude that some visual similarity facilitates transfer learning from the pretraining dataset to the target dataset. However, this is one of many aspects to consider, as we will see below.
Another visual aspect to consider is the average character width. As mentioned in Section~\ref{sec:implementation_details}, HWT generates images that are 32 pixels high and have an average character width of 16 pixels. This aspect ratio is very different from that of the images in the target datasets, which, on average, have a character width of around 8 pixels. For this reason, by using the few examples available, we estimated a form factor to shrink the width of the HWT-generated images to match those of the target datasets. The results of the CRNN models pretrained on the width-adjusted (WA) variant of the synthetics datasets, reported in Tables~\ref{tab:direct_transfer}-\ref{tab:saint_fine_tuning}, highlight that a smaller average character width, which is similar to the character width of the target datasets, leads to better performance compared to the original HWT-generated version.  

\tit{Handwriting}
\begin{table}[t]
\footnotesize
\centering
\caption{Performance of the considered model when pretrained on real datasets or on differently-obtained synthetic ones (WA stands for `width adjustment' and VS for `varied styles') and fine-tuned on different portions of the training set of Leopardi.}
\label{tab:leo_fine_tuning}
\setlength{\tabcolsep}{.4em}
\resizebox{\linewidth}{!}{
\begin{tabular}{lc cc c cc c cc c cc c cc}
\toprule
& & \multicolumn{14}{c}{\textbf{Fine-tuning/Training on}}\\
& & \multicolumn{2}{c}{\makecell{1.25\%\\(15 l.)}} & & \multicolumn{2}{c}{\makecell{2.5\%\\(32 l.)}} & & \multicolumn{2}{c}{\makecell{5\%\\(65 l.)}} & & \multicolumn{2}{c}{\makecell{50\%\\(652 l.)}} & & \multicolumn{2}{c}{\makecell{100\%\\(1303 l.)}} \\
\cmidrule{3-4} \cmidrule{6-7} \cmidrule{9-10} \cmidrule{12-13} \cmidrule{15-16} 
& & CER & WER & & CER & WER & & CER & WER & & CER & WER & & CER & WER \\
\midrule
Leopardi & & - & - & & - & - & & - & - & & 4.9 & 18.4 & & 2.8 & 10.8\\
\midrule
IAM & & 21.7 & 63.2 & & 17.1 & 54.1 & & 12.0 & 41.0 & & 4.9 & 19.4 & & 3.3 & 13.5 \\
RIMES & & 25.3 & 68.7 & & 18.2 & 54.6 & & 13.1 & 42.3 & & 4.3 & 15.7 & & 2.7 & 10.5 \\
Rodrigo & & 34.9 & 81.1 & & 23.0 & 63.7 & & 15.8 & 48.8 & & 5.2 & 18.6 & & 2.9 & 11.0 \\
ICFHR14 & & 23.6 & 67.7 & & 16.6 & 53.6 & & 11.7 & 40.7 & & 4.0 & 15.9 & & 2.7 & 10.8 \\
ICFHR16 & & 38.7 & 85.8 & & 24.4 & 68.1 & & 16.7 & 53.4 & & 6.3 & 24.2 & & 4.3 & 17.8 \\
NorHand & & 21.0 & 63.1 & & 15.2 & 50.3 & & 11.5 & 40.2 & & \textbf{3.4} & 13.2 & & \textbf{2.3} & 8.7 \\
LAM & & \textbf{12.7} & \textbf{42.0} & & \textbf{12.1} & \textbf{39.6} & & \textbf{8.2} & \textbf{28.8} & & 3.5 & \textbf{12.7} & & \textbf{2.3} & \textbf{8.6} \\
\midrule
Human-Synthesized  & & 25.3 & 70.1 & & 17.5 & 53.8 & & 11.7 & 38.5 & & 4.3 & 16.1 & & 2.5 & 9.6 \\ 
HWT-Generated  & & 66.3 & 97.8 & & 43.8 & 84.8 & & 22.5 & 59.6 & & 5.0 & 18.3 & & 2.8 & 10.9 \\ 
HWT-Generated+WA  & & 43.4 & 89.2 & & 27.9 & 70.2 & & 17.4 & 51.4 & & 5.1 & 18.6 & & 2.7 & 10.3 \\ 
HWT-Generated+WA+VS & & 35.6 & 80.9 & & 23.6 & 63.8 & & 14.8 & 45.6 & & 4.5 & 16.9 & & 2.6 & 10.2 \\ 
\bottomrule
\end{tabular}
}
\vspace{-0.3cm}
\end{table}
\begin{table}[t]
\footnotesize
\centering
\caption{Performance of the considered model when pretrained on real datasets or on differently-obtained synthetic ones (WA stands for `width adjustment' and VS for `varied styles') and fine-tuned on different portions of the training set of Washington.}
\label{tab:wash_fine_tuning}
\setlength{\tabcolsep}{.4em}
\resizebox{\linewidth}{!}{
\begin{tabular}{lc cc c cc c cc c cc c cc}
\toprule
& & \multicolumn{14}{c}{\textbf{Fine-tuning/Training on}}\\
& & \multicolumn{2}{c}{\makecell{1.25\%\\(6 l.)}} & & \multicolumn{2}{c}{\makecell{2.5\%\\(13 l.)}} & & \multicolumn{2}{c}{\makecell{5\%\\(26 l.)}} & & \multicolumn{2}{c}{\makecell{50\%\\(263 l.)}} & & \multicolumn{2}{c}{\makecell{100\%\\(526 l.)}} \\
\cmidrule{3-4} \cmidrule{6-7} \cmidrule{9-10} \cmidrule{12-13} \cmidrule{15-16} 
& & CER & WER & & CER & WER & & CER & WER & & CER & WER & & CER & WER \\
\midrule
Washington & & - & - & & - & - & & - & - & & 5.3 & 24.3 & & 3.4 & 15.9 \\
\midrule
IAM & & \textbf{18.8} & \textbf{52.7} & & 14.4 & 45.9 & & 12.5 & 40.0 & & 4.9 & 20.7 & & 3.9 & 16.1 \\
RIMES & & 27.1 & 78.9 & & 20.8 & 65.2 & & 16.8 & 55.7 & & 4.6 & 20.7 & & 3.7 & 17.1 \\
Rodrigo & & 48.9 & 92.6 & & 39.6 & 84.3 & & 27.3 & 69.0 & & 6.4 & 25.8 & & 4.5 & 19.9 \\
ICFHR14 & & 26.1 & 64.6 & & \textbf{14.2} & \textbf{43.9} & & \textbf{11.1} & \textbf{36.2} & & \textbf{3.9} & \textbf{17.7} & & \textbf{2.8} & \textbf{12.9} \\
ICFHR16 & & 58.0 & 97.8 & & 49.1 & 90.1 & & 30.3 & 74.6 & & 7.1 & 28.0 & & 5.2 & 21.3 \\
NorHand & & 31.4 & 76.7 & & 21.7 & 61.2 & & 16.1 & 52.9 & & 5.8 & 23.7 & & 3.7 & 15.3 \\
LAM & & 37.2 & 81.7 & & 27.6 & 72.6 & & 20.7 & 60.7 & & 5.9 & 23.1 & & 4.8 & 20.3 \\
\midrule
HWT-Generated  & & 54.9 & 93.8 & & 43.4 & 84.7 & & 28.8 & 70.8 & & 5.3 & 20.9 & & 3.7 & 16.5 \\
HWT-Generated+WA  & & 47.4 & 90.9 & & 34.0 & 79.7 & & 27.9 & 70.8 & & 5.5 & 22.1 & & 3.6 & 17.3 \\
HWT-Generated+WA+VS & & 31.3 & 77.7 & & 23.9 & 63.4 & & 19.1 & 56.9 & & 4.8 & 20.9 & & 3.3 & 16.1 \\
\bottomrule
\end{tabular}
}
\vspace{-0.3cm}
\end{table}
\begin{table}[h!]
\footnotesize
\centering
\caption{Performance of the considered model when pretrained on real datasets or on differently-obtained synthetic ones (WA stands for `width adjustment' and VS for `varied styles') and fine-tuned on different portions of the training set of Saint Gall.}
\label{tab:saint_fine_tuning}
\setlength{\tabcolsep}{.4em}
\resizebox{\linewidth}{!}{
\begin{tabular}{lc cc c cc c cc c cc c cc}
\toprule
& & \multicolumn{14}{c}{\textbf{Fine-tuning/Training on}}\\
& & \multicolumn{2}{c}{\makecell{1.25\%\\(5 l.)}} & & \multicolumn{2}{c}{\makecell{2.5\%\\(11 l.)}} & & \multicolumn{2}{c}{\makecell{5\%\\(23 l.)}} & & \multicolumn{2}{c}{\makecell{50\%\\(234 l.)}} & & \multicolumn{2}{c}{\makecell{100\%\\(468 l.)}} \\
\cmidrule{3-4} \cmidrule{6-7} \cmidrule{9-10} \cmidrule{12-13} \cmidrule{15-16} 
& & CER & WER & & CER & WER & & CER & WER & & CER & WER & & CER & WER \\
\midrule
Saint Gall & & - & - & & - & - & & - & - & & 5.8 & 38.6 & & 4.5 & 32.5 \\
\midrule
IAM & & 16.5 & 68.3 & & 13.3 & 61.0 & & 10.6 & 54.2 & & 5.4 & 36.0 & & 4.6 & 31.4 \\
RIMES & & 28.2 & 94.2 & & 19.7 & 79.7 & & 14.3 & 66.8 & & 6.5 & 39.9 & & 5.8 & 36.9 \\
Rodrigo & & \textbf{14.4} & \textbf{66.4} & & \textbf{11.3} & \textbf{58.7} & & \textbf{8.8} & \textbf{50.8} & & \textbf{5.3} & 35.8 & & 4.6 & 31.9 \\
ICFHR14 & & 20.4 & 77.8 & & 16.6 & 70.5 & & 12.1 & 54.2 & & \textbf{5.3} & 35.4 & & \textbf{4.5} & \textbf{30.9} \\
ICFHR16 & & 32.0 & 94.2 & & 22.8 & 83.5 & & 16.1 & 71.1 & & 6.8 & 41.9 & & 5.7 & 36.0 \\
NorHand & & 27.0 & 87.9 & & 19.5 & 75.0 & & 12.8 & 61.3 & & \textbf{5.3} & \textbf{35.1} & & 4.6 & 31.5 \\
LAM & & 20.8 & 77.8 & & 15.8 & 68.1 & & 12.2 & 60.0 & & \textbf{5.3} & 35.5 & & 4.6 & 31.4 \\
\midrule
HWT-Generated  & & 20.4 & 77.8 & & 16.6 & 70.5 & & 12.1 & 58.8 & & 5.5 & 37.5 & & 4.8 & 33.3 \\
HWT-Generated+WA  & & 19.7 & 80.1 & & 14.3 & 66.6 & & 11.4 & 58.6 & & 5.4 & 36.8 & & \textbf{4.5} & 31.2 \\
HWT-Generated+WA+VS & & 18.8 & 76.1 & & 13.3 & 61.2 & & 11.0 & 55.8 & & 5.4 & 35.9 & & \textbf{4.5} & 31.6 \\
\bottomrule
\end{tabular}
}
\vspace{-0.3cm}
\end{table}
By observing the results in Tables~\ref{tab:direct_transfer}-\ref{tab:saint_fine_tuning} alongside the datasets information in Table~\ref{tab:datasets} (especially the number of authors and the time-span), we can highlight a correlation between the performance and the different calligraphies in the pertaining dataset. This latter, in particular, often ensures a high variability of the handwriting style of the images in the dataset.
If the pretraining dataset contains texts written by multiple authors, the network is exposed to high variance and will achieve the ability to handle different styles and calligraphies. On the other hand, pretraining the network on images with only one author's handwriting reduces the variance and makes the network focus on that single author. 
For example, the LAM dataset has low variance since it is single-author. Nonetheless, it is similar to the Leopardi dataset due to language similarities and the historical period. Thus, pretraining on LAM induces a bias that allows the HTR model to effectively generalize to Leopardi. 
An example of the opposite case can be observed when pretraining on the ICFHR16 dataset and fine-tuning on Saint Gall (Table~\ref{tab:saintVSother}). ICFHR16 is a German dataset with a significant difference compared to Saint Gall, which is in Latin. Moreover, since ICFHR16 contains texts written by multiple authors, the dataset has a higher variance than Saint Gall, which is single-author. Therefore, during fine-tuning, the network needs to apply more corrections to adjust for the bias and reduce the variance to focus on the Saint Gall texts, and therefore, more samples are needed to achieve good results.
Overall, the results in Tables~\ref{tab:direct_transfer},~\ref{tab:leo_fine_tuning},~\ref{tab:saint_fine_tuning}, and~\ref{tab:wash_fine_tuning} show that, on average, all single-author datasets bring to better performance on Leopardi and Saint Gall, while in the Washington dataset, which contains the texts written by two authors, the multi-author datasets  (\eg~the IAM dataset) with many different styles with a high variance allow obtaining better results.

\tit{Language}
\begin{table}[t]
\centering
\small
\setlength{\tabcolsep}{.3em}
\caption{Language comparison between the Leopardi dataset and the considered pertaining datasets, ordered by average Kullback Leiber Divergence of n-grams.}
\label{tab:leoVSother}
\resizebox{0.85\columnwidth}{!}{%
\begin{tabular}{l c c c ccc c cc}
\toprule 
& & \multirow{2}{1.8cm}{\hfil\textbf{Lexical Similarity}} & & \multicolumn{3}{c}{\textbf{Kullback Leiber Divergence}} & & \multicolumn{2}{c}{\textbf{FT on 1.25\%}}\\
 \cmidrule{5-7} \cmidrule{9-10}
 & & & & {Unigram} & {Bigram} & {Trigram} & & {CER} & {WER}\\
\midrule
LAM          & & - & & 0.02 & 0.09 & 0.23 & & 12.7 & 42.0 \\
Synthetic for Leopardi     & & - & & 0.05 & 0.19 & 0.54 & & 35.6 & 80.9 \\
RIMES        & & 9.54 & & 0.11 & 0.84 & 1.89 & & 25.3 & 68.7 \\
IAM          & & 6.76 & & 0.17 & 0.85 & 1.62 & & 21.7 & 63.2 \\
ICFHR14      & & 6.76 & & 0.17 & 0.91 & 1.83 & & 23.6 & 67.7 \\
Rodrigo      & & 10.45 & & 0.20 & 0.71 & 1.48 & & 34.9 & 81.1 \\
NorHand  & & 3.99 & & 0.34 & 1.15 & 2.06 & & 21.0 & 63.1 \\
ICFHR16      & & 4.19 & & 0.40 & 1.50 & 2.50 & & 39.0 & 86.0 \\
\bottomrule
\end{tabular}
}\\
\vspace{-0.3cm}
\end{table}
\begin{table}[t]
\centering
\small
\setlength{\tabcolsep}{.3em}
\caption{Language comparison between the Saint Gall dataset and the considered pertaining datasets, ordered by average Kullback Leiber Divergence of n-grams.}
\label{tab:saintVSother}
\resizebox{0.85\columnwidth}{!}{%
\begin{tabular}{l c c c ccc c cc}
\toprule 
& & \multirow{2}{1.8cm}{\hfil\textbf{Lexical Similarity}} & & \multicolumn{3}{c}{\textbf{Kullback Leiber Divergence}} & & \multicolumn{2}{c}{\textbf{FT on 1.25\%}}\\
 \cmidrule{5-7} \cmidrule{9-10}
 & & & & {Unigram} & {Bigram} & {Trigram} & & {CER} & {WER}\\
\midrule
LAM          & & 5.81 & & 0.17 & 0.87 & 1.59 & & 20.8 & 77.8 \\
Rodrigo      & & 6.08 & & 0.18 & 0.89 & 1.74 & & 14.4 & 66.4 \\
RIMES        & & 5.39 & & 0.19 & 0.87 & 1.74 & & 28.2 & 94.2 \\
ICFHR14      & & 3.50 & & 0.20 & 0.79 & 1.43 & & 20.4 & 77.8 \\
IAM          & & 3.50 & & 0.21 & 0.74 & 1.31 & & 16.5 & 68.3 \\
Synthetic for Saint Gall   & & - & & 0.23 & 0.60 & 1.08 & & 18.8 & 76.1 \\
NorHand  & & 2.39 & & 0.42 & 1.22 & 1.78 & & 27.0 & 87.9 \\
ICFHR16      & & 2.73 & & 0.58 & 1.60 & 2.10 & & 32.0 & 94.2 \\
\bottomrule
\end{tabular}
}\\
\vspace{-.3cm}
\end{table}
\begin{table}[t]
\centering
\small
\setlength{\tabcolsep}{.3em}
\caption{Language comparison between the Washington dataset and the considered pertaining datasets, ordered by average Kullback Leiber Divergence of n-grams.}
\label{tab:washVSother}
\resizebox{0.85\columnwidth}{!}{%
\begin{tabular}{l c c c ccc c cc}
\toprule 
& & \multirow{2}{1.8cm}{\hfil\textbf{Lexical Similarity}} & & \multicolumn{3}{c}{\textbf{Kullback Leiber Divergence}} & & \multicolumn{2}{c}{\textbf{FT on 1.25\%}}\\
 \cmidrule{5-7} \cmidrule{9-10}
 & & & & {Unigram} & {Bigram} & {Trigram} & & {CER} & {WER}\\
\midrule
ICFHR14      & & - & & 0.05 & 0.30 & 0.66 & & 26.1 & 64.6 \\
Synthetic for Washington     & & - & & 0.07 & 0.31 & 0.69 & & 23.9 & 63.4 \\
IAM          & & - & & 0.08 & 0.30 & 0.59 & & 18.8 & 52.7 \\
NorHand  & & 4.30 & & 0.29 & 1.03 & 1.64 & & 31.4 & 76.7 \\
RIMES        & & 9.67 & & 0.31 & 1.36 & 2.22 & & 27.1 & 78.9 \\
Rodrigo      & & 7.91 & & 0.35 & 1.38 & 2.32 & & 48.9 & 92.6 \\
ICFHR16      & & 4.72 & & 0.36 & 1.24 & 1.83 & & 58.0 & 97.8 \\
LAM          & & 6.76 & & 0.36 & 1.52 & 2.31 & & 37.2 & 81.7 \\
\bottomrule
\end{tabular}
}\\
\vspace{-.3cm}
\end{table}
Tables~\ref{tab:leoVSother},~\ref{tab:saintVSother}, and~\ref{tab:washVSother} compare the language similarity and the Kullback Leiber Divergence (KL) between the pretraining and target datasets. To highlight the correlation between the textual information and the network performance, we sort the tables by KL divergence and include the results obtained after fine-tuning with the $1.25\%$ of the target dataset. In this way, we emphasize a trend where datasets with a slight language variation are more suitable to be used in pretraining than datasets with a significant difference. In particular, ICFHR16, in German, is, on average, the farthest dataset to all the target datasets we compare with in terms of KL. As a result, pretraining on this dataset leads to the highest recognition errors in all three target datasets. On the other hand, pretraining on IAM and ICFHR14, which are in English, allows obtaining good performance thanks to the language similarity to all the target datasets.  
Notably, from Table~\ref{tab:leoVSother}, we observe that from a lexical point of view, the Leopardi dataset is closer to the LAM dataset than the synthetic one containing proses by the author. The text in all three datasets is in Italian and was written in the same period with a time difference of fewer than 70 years. Arguably, the reason why LAM is closer to the Leopardi dataset is that both datasets are a collection of letters, which share many structural similarities (\eg~dates, openings, salutations). Combining this aspect, the language, and the period makes CRNN pretrained on the LAM dataset to obtain impressive performance on the Leopardi test set, particularly in a direct transfer setting (see Table~\ref{tab:direct_transfer}).
\section{Conclusion}\label{sec:conclusion}
In this paper, we have explored line-level HTR on historical manuscripts when limited training data are available. To this end, we have proposed to pretrain a dedicated HTR model on existing benchmark datasets or on a large quantity of synthetic data that reflect the characteristics of the handwriting of the target author of the manuscripts, which we built with a fully-automatic procedure, and fine-tune on a portion of real data in the collection of interest. In particular, we have conducted an extensive quantitative analysis of the main characteristics that the pretraining dataset should have in order to obtain a strong HTR model with as little as five lines from the target collection. 

The obtained experimental results show that when choosing the real dataset or generating the synthetic one for pretraining, both the overall appearance (given by the paper support, writing tool, and average character width) and the language should be taken into account. Moreover, it has emerged that an HTR model trained on images of text with high variability in handwriting style is more robust and easily adaptable than one trained on a single handwriting style. Nonetheless, when the synthetic data faithfully resemble the real ones in terms of handwriting, satisfactory performance is achievable. 

In the sight of these conclusions, this work can help guide the selection of the most suitable pretraining dataset to boost the performance of HTR models on small domain-specific documents and give some insights into the maturity of the HTG field and its potential benefit for HTR.
Finally, this work has shed some light on the feasibility of interactive, on-demand HTR on single-author collections, which is a task worthy of further investigation for its application to digital archives use and enhancement.

\section*{Acknowledgement}
This work was supported by the ``AI for Digital Humanities'' project (Pratica Sime n.2018.0390), funded by ``Fondazione di Modena'' and the PNRR project Italian Strengthening of ESFRI RI Resilience (ITSERR) funded by the European Union – NextGenerationEU (CUP: B53C22001770006).

\clearpage

\bibliographystyle{splncs04}
\bibliography{bibliography}
\end{document}